%% file: iclr2026_conference.tex
\documentclass{article} 
\usepackage{iclr2026_conference,times}

\input{math_commands.tex}

\usepackage{hyperref}
\usepackage{url}
\usepackage{color}

\usepackage{booktabs}       
\usepackage{threeparttable} 
\usepackage{multirow}       
\usepackage{siunitx}    

\usepackage{graphicx}
\usepackage{algorithm}
\usepackage{algpseudocode}
\usepackage{amsmath}
\usepackage{booktabs}
\usepackage{caption}
\usepackage{listings}
\usepackage{amssymb}

\usepackage{tikz}
\usetikzlibrary{shapes,arrows.meta,positioning,fit,calc,backgrounds,shadows}

\title{VTPerception-R1: Enhancing Multimodal Reasoning via Explicit Visual and Textual Perceptual Grounding}

\author{Yizhuo Ding\textsuperscript{1,2}, Mingkang Chen\textsuperscript{2,3}, Zhibang Feng\textsuperscript{4}, Tong Xiao\textsuperscript{5}, Wanying Qu\textsuperscript{1,2}, \\
\textbf{Wenqi Shao\textsuperscript{2}, Yanwei Fu\textsuperscript{1}}\\
 \textsuperscript{1}Fudan University, \textsuperscript{2}Shanghai AI Laboratory, \textsuperscript{3}The University of Hong Kong\\ \textsuperscript{4}Shenzhen University, \textsuperscript{5}University of Science and Technology of China \\
 \texttt{\{yzding22\}@m.fudan.edu.cn};\\
  \texttt{\{yanweifu\}@fudan.edu.cn}; \texttt{\{shaowenqi\}@pjlab.org.cn} \\
 }

\newcommand{\modelname}{VTPerception-R1-7B}
\newcommand{\methodname}{VTPerception-R1}

\begin{document}

\maketitle
\input{chap/abstract}
\input{chap/introduction}

\input{chap/related_work}
\input{chap/verification}

\input{chap/method}
\input{chap/experiments}
\input{chap/conclusion}

\bibliography{iclr2026_conference}
\bibliographystyle{iclr2026_conference}

\appendix
\input{chap/appendix}
\end{document}

%% file: math_commands.tex
\usepackage{amsmath,amsfonts,bm}

\def\eqref#1{equation~\ref{#1}}

\def\1{\bm{1}}

\DeclareMathAlphabet{\mathsfit}{\encodingdefault}{\sfdefault}{m}{sl}
\SetMathAlphabet{\mathsfit}{bold}{\encodingdefault}{\sfdefault}{bx}{n}



%% file: chap/abstract.tex
\begin{abstract}

Multimodal large language models (MLLMs) often struggle to ground reasoning in perceptual evidence. We present a systematic study of perception strategies—explicit, implicit, visual, and textual—across four multimodal benchmarks and two MLLMs. Our findings show that explicit perception, especially when paired with textual cues, consistently yields the best improvements, particularly for smaller models. Based on this insight, we propose \methodname, a unified two-stage framework that decouples perception from reasoning. Stage I introduces perception-augmented fine-tuning, and Stage II applies perception-aware reinforcement learning with novel visual, textual, and consistency rewards. Experiments demonstrate that \methodname\ significantly improves reasoning accuracy and robustness across diverse tasks, offering a scalable and auditable solution for perception-grounded multimodal reasoning. Our code is available at: https://github.com/yizhuoDi/VTPerceprion-R1.
\end{abstract}

%% file: chap/introduction.tex
\section{Introduction}

Multimodal reasoning plays a crucial role in real-world applications such as  solving geometry problems and understanding complex image–text inputs. Recent developments have extended reinforcement learning with verifiable rewards (RLVR)~\citep{dong2024rlhf} from text-only large language models (LLMs) to multimodal large language models (MLLMs). When combined with GRPO-style~\citep{grpo} objectives, structured format and answer rewards, and improved data or rollout strategies, these methods achieve consistent performance gains. For example, MM-Eureka~\citep{mm_eureka} and R1-VL~\citep{zhang2025r1} demonstrate how RLVR can be stabilized in vision–language settings, while R1-OneVision and Vision-R1 adopt a “cold-start + RLVR” approach to bridge the modality gap and further enhance reasoning capabilities. However, these advances largely overlook a critical component: \textit{perceptual grounding}.

Perceptual grounding, whether implicit or explicit, has been shown to influence reasoning profoundly. PAPO~\citep{PAPO} introduces masked-image sensitivity with regularizers to guide reasoning implicitly. In contrast, Perception-R1~\citep{PerceptionR1} employs explicit perception by leveraging annotated feedback and external LLM-based judges. Open Vision Reasoner~\citep{wei2025open} introduces a two-stage pipeline that transfers linguistic reasoning patterns to the visual domain, showing that RL can compensate for the performance degradation of cold-start visual grounding. Unfortunately, despite advances in multimodal large language models (MLLMs), existing systems remain brittle: they struggle to ground reasoning in perceptual evidence, are prone to modality bias, and lack systematic methods to unify perception and reasoning. This limits their ability to tackle tasks requiring deep understanding of complex visual-textual contexts, such as interpreting scientific figures, solving geometry problems, or answering open-ended multimodal queries. Addressing these gaps is not only scientifically compelling but also critical for building robust AI capable of reliable, explainable decision-making in real-world environments.

Other methods take a more direct route by enforcing explicit perception through model outputs. Visionary-R1~\citep{xia2025visionary} imposes a “caption → reason → answer” structure, rewarding accurate captions that enable reasoning without shortcuts. Vision-SR1~\citep{li2025self} decomposes model outputs into visual descriptions, reasoning steps, and answers, and verifies the sufficiency of self-generated perceptions by re-prompting without images. This enables the use of self-visual rewards.

While both implicit and explicit approaches have shown benefits, most prior work lacks rigorous experimental verification to quantify the impact of visual perception—particularly in the case of explicit perception. PAPO conducts a manual audit revealing that 67\% of GRPO’s errors stem from perception issues, and shows that introducing an implicit perception KL reward reduces such errors by 30.5\%. Vision-SR1 performs an ablation study that removes the perception self-reward, resulting in a consistent performance drop. However, neither work conducts comprehensive experiments to systematically evaluate the effectiveness of explicit visual perception on reasoning and general benchmarks. Moreover, prior research~\citep{tourimpampa2018perception} highlights that both visual and textual perceptual cues jointly influence question comprehension. Yet, existing methods rarely incorporate textual perception, leading to potential over-reliance on visual inputs and increasing the risk of hallucinated reasoning.

This paper addresses these gaps through two core contributions. First, we conduct a systematic study of perception strategies for multimodal reasoning. We benchmark implicit and explicit visual and textual perception across four challenging datasets—\textbf{MMMU}, \textbf{MathVista}, \textbf{EMMA}, and a curated subset of \textbf{OlympiaBench}—using two MLLMs: \textbf{Qwen2.5-VL-32B} and \textbf{Qwen2.5-VL-7B}. We compare three grounding strategies: explicit perception notes, structured grounding via self-description, and implicit “look carefully” prompts. Our findings reveal that explicit perception consistently yields the largest performance gains, particularly for smaller models. We further show that textual perception provides meaningful improvements when capacity is limited, confirming that perceptual grounding is essential for robust multimodal reasoning and that vision and language cues jointly enhance model comprehension.

Second, we propose \methodname, a unified perception-grounded training framework that explicitly decouples seeing from reasoning. \methodname\ adopts a two-stage pipeline.

\noindent \textbf{i)Stage I (Perception-augmented SFT)} trains the model to produce concise, task-relevant \textless description\textgreater\ before reasoning and answering, establishing perceptual grounding as an integral part of the reasoning process.

\noindent \textbf{ii)Stage II (Perception-aware RL)} builds upon a DAPO-style objective and introduces novel perception-specific rewards: a visual perception reward ($R_{\text{vkey}}$) to measure coverage of key visual cues, a textual perception reward ($R_{\text{tkey}}$) to measure coverage of salient textual cues, and a consistency reward ($R_{\text{cons}}$) to ensure that reasoning and answers are grounded in the perceived evidence rather than hallucinations. A perception-first weighting schedule emphasizes perceptual grounding early in training before shifting focus to correctness, producing an auditable, balanced, and end-to-end SFT→RL framework. This approach contrasts with prior methods that rely on implicit constraints or modality-specific designs, offering a principled way to jointly optimize vision and language perception for multimodal reasoning.

Through extensive experiments, we demonstrate that \methodname\ significantly improves performance across diverse reasoning benchmarks, validating both the necessity of systematic perception grounding and the efficacy of our proposed framework. This work delivers the first comprehensive analysis of perception strategies for RLVR in MLLMs and provides a practical, scalable framework that bridges the gap between perception and reasoning.

This paper makes three key contributions:

\noindent \textit{(1) Systematic evaluation of perception strategies}. We conduct the first large-scale study comparing implicit and explicit visual/textual grounding across four multimodal benchmarks—\textbf{MMMU}, \textbf{MathVista}, \textbf{EMMA}, and a curated subset of \textbf{OlympiaBench}—using \textbf{Qwen2.5-VL-32B} and \textbf{Qwen2.5-VL-7B}. Results show explicit grounding consistently delivers the largest gains, especially for smaller models, with textual perception providing complementary improvements.

\noindent \textit{(2) \methodname}: A unified perception-grounded training framework. We propose \methodname, a two-stage framework that explicitly decouples seeing from reasoning. Stage I trains models to produce concise perceptual descriptions before reasoning, while Stage II applies perception-aware RL with novel visual, textual, and consistency rewards. A perception-first weighting schedule ensures balanced grounding and correctness.

\noindent \textit{(3)  Empirical validation and insights}. Extensive experiments demonstrate \methodname’s superior reasoning performance and robustness. Our results highlight the critical role of balanced visual and textual grounding, establishing \methodname\ as a principled framework for perception-grounded reasoning in MLLMs.

%% file: chap/related_work.tex
\section{Related Work}

\noindent \textbf{Large Multimodal Reasoning Models.}
Early work in multimodal reasoning combined prompt design with supervised chain-of-thought (CoT) to integrate vision and language. More recent approaches~\citep{grpo,ppo,dapo} adopt reinforcement learning with verifiable rewards (RLVR), optimizing both answer correctness and structured reasoning formats under GRPO/DAPO objectives. However, directly applying text-only RLVR to vision–language tasks exposes a perception bottleneck: many reasoning failures originate in the perceptual stage, demonstrating that correctness-only objectives are insufficient.

To address this, recent works strengthen perception–reasoning coupling. PAPO~\citep{PAPO} uses a reverse-KL objective and entropy regularization to improve perception awareness, reducing perception-related errors. R1-style pipelines~\citep{lmmr1,visionr1,reasonrft,mm_eureka,vl_rethinker} adapt text RLVR to MLLMs via two-stage training, reasoning control, curated datasets, and selective replay. Complementary work~\citep{vapo,dr_grpo,noisyrollout,r1_sharevl,skywork_r1v2} refines GRPO and explores rollout/data design. These approaches show that integrating perception with reasoning is more effective than optimizing correctness alone.

\noindent \textbf{Perception in Multimodal Models.}
Empirical studies show correctness-driven RLVR often fails to improve perception, with perception errors dominating failure cases~\citep{PerceptionR1,mathvista,mathverse}. Perception-targeted objectives address this: Perception-R1~\citep{PerceptionR1} rewards consistency between reasoning and visual annotations; PAPO~\citep{PAPO} couples perception and policy learning via reverse-KL and entropy regularization; SRPO~\citep{SRPO} improves consolidation of visual/textual cues to strengthen reasoning. Other works treat perception as tool use~\citep{deep_eyes,active_o3,pixel_reasoner}, enhancing external visual operations rather than intrinsic grounding. These lines show that explicit perception rewards and perception-aware objectives improve evidence coverage and reasoning quality.

%% file: chap/verification.tex
\section{Systematic  Study of Perception Strategies}
\label{sec:pre_val}

In this section, we systematically investigate how visual and textual perception influence reasoning performance. Our experiments are designed to quantify the effectiveness of different perception strategies and their impact on downstream multimodal reasoning tasks.
Settings

\subsection{Settings}

To assess the influence of visual and textual perception capabilities, we conducted controlled experiments across four diverse benchmarks: \textbf{MMMU}, \textbf{MathVista}, \textbf{EMMA}, and \textbf{OlympiaBench}. These benchmarks encompass varied reasoning challenges, enabling comprehensive evaluation of perception strategies.

For visual perception, we considered three experimental conditions. In the first condition,\textbf{explicit visual perception integration}, image-based queries were enriched with structured visual perception annotations appended to the input. This allows the model to leverage explicit perceptual representations, and performance was benchmarked against these enriched inputs. The second condition, \textbf{structured visual grounding prompting}, involved engineering the system prompt to explicitly require the model to articulate its perceptual analysis of the visual information prior to producing a final reasoning output. In the third condition, \textbf{implicit visual grounding prompting}, the system prompt was minimally modified to instruct the model to “carefully observe the image” before responding, priming visual attention without requiring explicit perceptual articulation.

For textual perception, we adopted analogous conditions. In the \textbf{structured visual–textual grounding prompting} condition, the model was prompted to articulate its integrated understanding of both visual and textual information before producing an answer, extending structured grounding to multimodal perception. In the \textbf{implicit visual–textual grounding prompting} condition, the system prompt was lightly adjusted to instruct the model to ``carefully observe the image and text'', encouraging implicit multimodal attention without explicit grounding outputs.

These settings enable a rigorous, controlled comparison of perception strategies, isolating their contributions to reasoning performance across diverse tasks. For all evaluations, we use the full datasets unless otherwise specified. Specifically, for OlympiaBench, our analysis was conducted on a carefully selected subset of 600 instances, sampled randomly and proportionally across all subtasks to ensure a representative distribution of the benchmark.

\subsection{Results and Lessons}

Table~\ref{tab:main_results} shows performance differences across system prompts and explicit data integration.

\begin{table}
\centering
\small
\setlength{\tabcolsep}{6pt}
\resizebox{1.0\textwidth}{!}{
\begin{tabular}{llcccc}
\toprule
\textbf{Model} & \textbf{Perception Setting} & \textbf{MathVista} & \textbf{MMMU} & \textbf{EMMA} & \textbf{OlympiaBench} \\
\midrule
\multirow{6}{*}{Qwen2.5-VL-32B}
& Baseline &71.11 & 60.57 & 30.85 & 16.73 \\
\cmidrule(lr){2-6}
& Explicit visual notes        & 73.33 & 62.33 & 31.97 & 18.44 \\
\cmidrule(lr){2-6}
& Structured visual grounding                  & 71.39 & 61.33 & 31.71 & 16.71 \\
& Structured visual--text grounding            & 71.41 & 61.28 & 31.67 & 16.76 \\
\cmidrule(lr){2-6}
& Implicit visual grounding                    & 70.40 & 60.67 & 30.43 & 17.85 \\
& Implicit visual--text grounding              & 70.57 & 60.42 & 30.67 & 17.88 \\
\midrule
\multirow{6}{*}{Qwen2.5-VL-7B}
& Baseline &61.67 & 46.91 &27.67 &7.03 \\
\cmidrule(lr){2-6}
& Explicit visual notes        & 62.30 & 48.28 & 28.36 & 7.60 \\
\cmidrule(lr){2-6}
& Structured visual grounding                  & 62.00 & 44.00 & 27.58 & 6.20 \\
& Structured visual--text grounding            & 62.30 & 45.11 & 27.87 & 6.42 \\
\cmidrule(lr){2-6}
& Implicit visual grounding                    & 61.90 & 47.67 & 27.97 & 7.20 \\
& Implicit visual--text grounding              & 62.30 & 48.11 & 28.33 & 7.11 \\
\bottomrule
\end{tabular}
}
\vspace{-0.1in}
\caption{Combined evaluation of Qwen2.5-VL-32B and Qwen2.5-VL-7B under visual and textual perception settings. Higher is better.\label{tab:main_results} }
\vspace{-0.1in}
\end{table}

\noindent \textbf{Lessons on Visual Perception}. 

\noindent \textit{(1) Qwen2.5VL-32B} Direct input augmentation with visual annotations achieves the highest overall performance, demonstrating the clear advantage of providing explicit visual understanding. Prompting the model to generate its own visual interpretation is moderately effective but generally less robust than supplying pre-processed annotations. Notably, on complex datasets such as OlympiaBench, structured prompting can be detrimental: when the model’s intrinsic perceptual ability is insufficient, requiring it to articulate its perception often produces hallucinated or inaccurate observations, introducing bias and degrading reasoning performance.

\noindent \textit{(2) Qwen2.5VL-7B} Similar to the 32B model, explicit visual annotation improves performance. However, the 7B model shows greater sensitivity to perception prompting, with structured prompts consistently reducing performance, particularly on more challenging tasks. This suggests that smaller models are more prone to self-induced perceptual errors when required to generate their own interpretations.

\noindent \textbf{Lessons on Textual Perception}.

\noindent \textit{(1)Qwen2.5-VL-32B.}
Under both explicit and implicit visual perception settings, incorporating additional textual perception yields only marginal performance gains. Moreover, the improvements are inconsistent across different benchmarks.

\noindent \textit{(2)Qwen2.5-VL-7B.}
Similar to the 32B variant, introducing textual perception on top of either explicit or implicit visual prompts results in limited performance improvement. However, the gains are notably more stable across tasks.

\textbf{Overall}. 
Perception prompting’s impact strongly depends on model scale: larger models leverage perceptual signals more effectively, while smaller models often struggle without explicit guidance. Supplying robust perceptual information boosts performance, confirming that strong perception is critical for advanced reasoning. Visual perception remains a major frontier for improvement across models, whereas textual perception offers particularly high gains for smaller-scale systems.

%% file: chap/method.tex
\section{Method}
\label{sec:method}

As discussed in Sec.~\ref{sec:pre_val}, our analysis reveals three key observations. First, both the 32B and 7B models benefit from explicit perception prompting, with the 7B model showing greater potential for improvement. Second, the 7B model’s baseline visual perception is relatively weak, requiring additional supervision to reliably ground visual information. Third, responses to textual perception prompts are unstable, indicating the need for better perception alignment.

\begin{figure}
    \centering
    \includegraphics[width=0.9\linewidth]{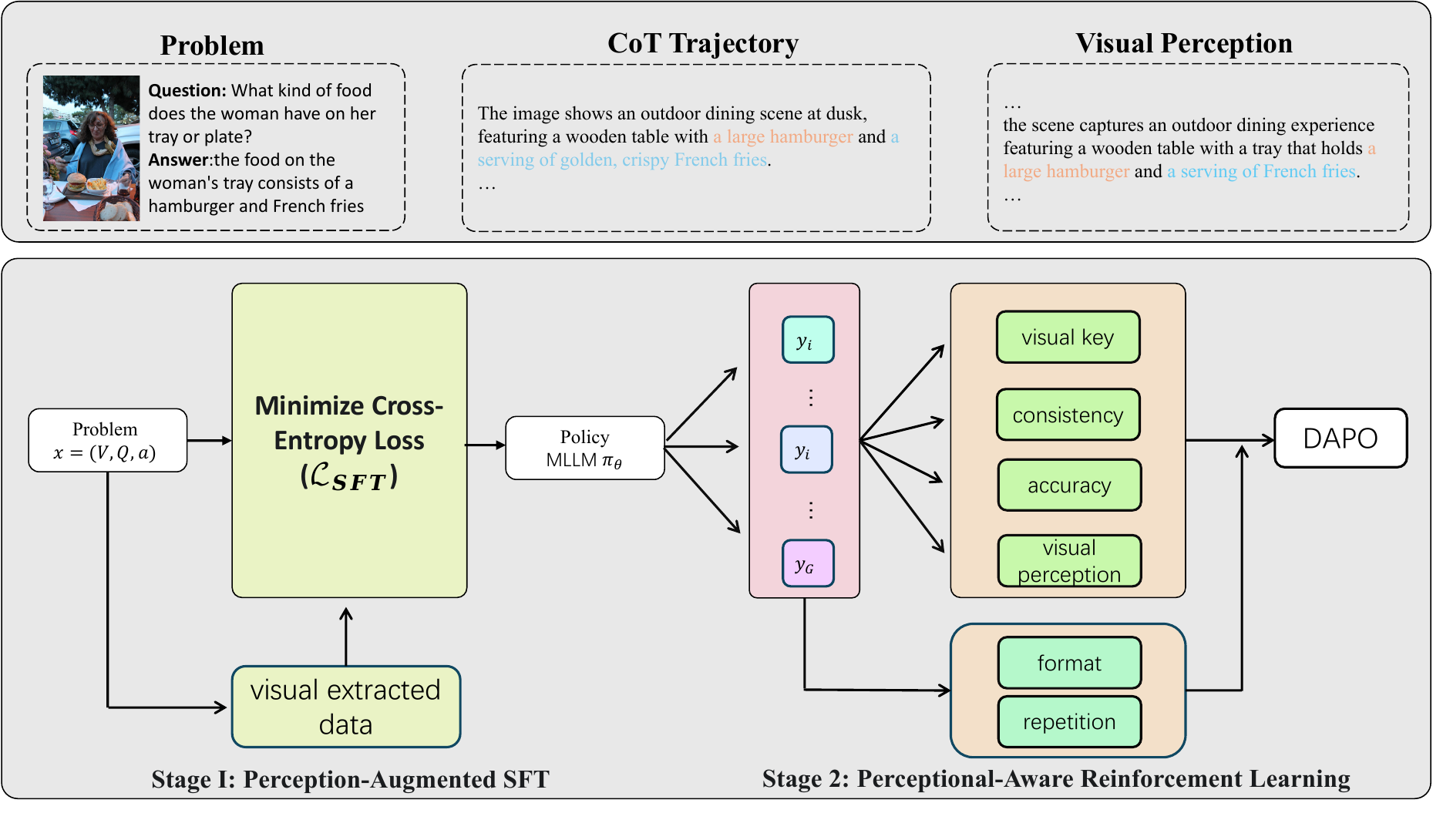}
\caption{Overview of the proposed two-stage training pipeline \methodname. 
Stage~1 performs supervised fine-tuning with perception-grounded annotations, 
where explicit visual and textual notes are integrated into the reasoning process to strengthen multimodal perception. 
Stage~2 applies reinforcement learning with perception-aware rewards, further refining the Description $\rightarrow$ Think $\rightarrow$ Answer reasoning pipeline for improved consistency and interpretability.}    
    \vspace{-0.1in}
    \label{fig:main_fig}
\end{figure}

Motivated by these insights, we propose a two-stage training pipeline, \methodname, to systematically enhance multimodal perception, particularly for the 7B variant. The overall framework is illustrated in Fig.~\ref{fig:main_fig}.

\noindent \textit{Stage 1}: Visual Perception–Augmented Supervised Fine-Tuning (SFT). This stage trains the model to generate focused <description> fields, improving its ability to extract and articulate visual content and grounding reasoning inputs.

\noindent \textit{Stage 2}: Perception-Aware Reinforcement Learning (RL). Building on Stage 1, this stage refines reasoning with a composite reward that encourages (i) capturing key evidence in \textless description\textgreater and \textless think\textgreater steps, and (ii) logical consistency between evidence and answers.

This design follows two principles: (a) Make perception explicit before reasoning to improve grounding and transparency, and (b) Reward what matters for reasoning to ensure faithful, consistent outputs. Empirically, \methodname significantly boosts evidence recall, reasoning consistency, and answer quality, especially for perception-deficient smaller models.

\subsection{Stage I: Perception-Augmented SFT}
\label{sec:sft}

\paragraph{Data and template.}
Starting from LLaVA-CoT style data, we convert each instance to a structured target:
\begin{equation}
\small
\textless description\textgreater \ldots \textless/description\textgreater
\textless think\textgreater \ldots \textless/think\textgreater
\textless answer\textgreater \ldots \textless/answer\textgreater.
\end{equation}
Here, \textless description\textgreater summarizes only the visual/textual evidence that is relevant to the question and useful for reasoning (not a generic caption). We retain the original chain-of-thought inside \textless think\textgreater and the final solution in \textless answer\textgreater. This explicit structuring encourages the model to first ground the input in perception, then conduct reasoning in a disentangled manner.

\paragraph{Objective.}

We minimize token-level cross-entropy over the full target sequence:
\begin{equation}
\mathcal{L}_{\text{SFT}} \;=\; - \sum_{t} \log \pi_{\theta}\!\big(y_t \,\big|\, x, y_{<t}\big),
\end{equation}
which bootstraps the model to (i) attend to question-relevant regions, (ii) verbalize them succinctly in \textless description\textgreater, and (iii) reason over these cues. This establishes a stable perception-to-description interface that Stage~II can further refine.

\paragraph{Motivation.}
Directly training on raw CoT-style data often entangles perception, reasoning, and answering, making it difficult for the model to learn robust visual grounding. By inserting the \textless description\textgreater stage, we encourage the model to first distill multimodal evidence into a compact, interpretable representation. This intermediate step reduces spurious correlations, improves generalization, and provides a controllable interface that can be explicitly inspected or modified during inference. In practice, the \textless description\textgreater field resembles a structured “scratchpad” for perception, while the subsequent \textless think\textgreater step focuses purely on abstract reasoning.

\paragraph{Effect.}
This stage produces a model that is able to (i) highlight salient objects and attributes, (ii) capture their spatial or semantic relations, and (iii) chain them into reasoning sequences leading to the final answer. Empirically, we find that Perception-Augmented SFT improves consistency between visual grounding and reasoning, and also simplifies downstream alignment, since the model is already trained to organize its outputs into modular components.

\subsection{Stage II: Perception-Aware RL}
\label{sec:rl}

\noindent \textbf{Decoupled Clip and Dynamic sAmpling Policy Optimization (DAPO)}.
DAPO is a reinforcement learning algorithm designed for large-scale reasoning models with long chain-of-thought outputs. Compared to earlier methods such as GRPO, it improves both stability and efficiency by introducing four key techniques: asymmetric clipping, dynamic sampling, token-level optimization, and overlong reward shaping.  Formally, for each prompt $x$, $G$ responses $\{o_i\}_{i=1}^G$ are sampled, and group-relative advantages are computed to normalize rewards. The training objective is a clipped-ratio policy gradient:
\[
J(\theta) \;=\; 
\mathbb{E}_{\{o_i\}\sim \pi_{\theta_{\text{old}}}}
\Bigg[
\frac{1}{\sum_i |o_i|}\sum_{i=1}^{G}\sum_{t=1}^{|o_i|}
\min\!\Big(
r_{i,t}(\theta)\,\widehat{A}_{i,t},\;
\mathrm{clip}\big(r_{i,t}(\theta),\,1-\varepsilon_{\text{low}},\,1+\varepsilon_{\text{high}}\big)\,\widehat{A}_{i,t}
\Big)
\Bigg],
\]
where $r_{i,t}(\theta)$ is the token-level importance ratio and $\widehat{A}_{i,t}$ the normalized advantage. To avoid degenerate updates, DAPO excludes trivial groups where all rollouts are correct or all are wrong.  

These improvements allow DAPO to provide stronger learning signals for long reasoning traces, improving convergence and reducing instability. In our work, DAPO serves as the backbone optimization method, enabling perception-aware rewards to directly shape both description quality and reasoning accuracy.

\noindent \textbf{Perception-Aware Reward}.
As is described in Section~\ref{sec:case_ver}, we motivated the model to enhance the capability of description before think answer. We want the model to reason according to the information they have seen. What is more, the combined attention of textual  information and visual information are all important. The total reward is a weighted sum:

\begin{equation}
\label{eq:total-reward}
R \;=\;  R_{\text{acc}}
\;+\;  R_{\text{fmt}}
\;+\;  R_{\text{vkey}}
\;+\;  R_{\text{tkey}}
\;+\; R_{\text{rep}}
\;+\;  R_{\text{cons}}.
\end{equation}
We describe each component below.

\noindent \textbf{(1) Answer accuracy ($R_{\text{acc}}$).}
This reward measures whether \textless answer\textgreater\ matches the ground truth, providing a direct, task-specific signal that reinforces correct reasoning and supports perception-related rewards.

\textbf{(2) Format compliance ($R_{\text{fmt}}$).}
This reward enforces the template \textless description\textgreater $\rightarrow$ \textless think\textgreater $\rightarrow$ \textless answer\textgreater, penalizing missing or repeated segments to ensure consistent, perception-grounded reasoning.

\textbf{(3) Repetition control ($R_{\text{rep}}$).}
This component penalizes repeated $n$-grams to discourage verbosity and promote concise, evidence-rich \textless description\textgreater\ and \textless think\textgreater\  segments.

\noindent \textbf{(4) Visual key-info ($R_{\text{vkey}}$).}
We enhance the data with key visual cues that are directly relevant to the task, ensuring the model captures meaningful visual information.
The reward measures how well \textless description\textgreater\ covers these critical visual elements. Given a key set $K$ and the set of facts $D$ extracted from \textless description\textgreater, we compute the recall $\text{cov}=\tfrac{|K\cap D|}{|K|}$ and discretize it into three levels:
\begin{equation}
R_{\text{vkey}} \;=\;
\begin{cases}
1.0, & \text{if } \text{cov} \ge \tau_{\text{hi}},\\
0.5, & \text{if } \tau_{\text{lo}} \le \text{cov} < \tau_{\text{hi}},\\
0.0, & \text{otherwise}.
\end{cases}
\end{equation}
This reward encourages the model to capture task-relevant visual evidence faithfully, thereby improving its ability to perceive and utilize visual information in downstream reasoning.

\noindent \textbf{(5) Textual key-info ($R_{\text{tkey}}$).}
To strengthen textual perception, we augment training data with task-specific key information—such as OCR text, numerical values, unit constraints, symbols, and relevant commonsense cues.
The reward measures how well \textless think\textgreater\ incorporates these critical textual elements. Given a key set $K$ and the set of facts $D$ extracted from \textless think\textgreater, we compute the recall $\text{cov}=\tfrac{|K\cap D|}{|K|}$ and discretize it into three levels:
\begin{equation}
R_{\text{tkey}} \;=\;
\begin{cases}
1.0, & \text{if } \text{cov} \ge \tau_{\text{hi}},\\
0.5, & \text{if } \tau_{\text{lo}} \le \text{cov} < \tau_{\text{hi}},\\
0.0, & \text{otherwise}.
\end{cases}
\end{equation}
As a complementary signal to $R_{\text{vkey}}$, this reward ensures that reasoning does not ignore textual evidence and helps maintain a balance between visual and textual perception.

\noindent \textbf{(6) Desc–reasoning consistency ($R_{\text{cons}}$).}
Finally, we add a consistency reward to align perception with reasoning. It checks whether the entities, attributes, and numerical values referenced in \textless think\textgreater\ and \textless answer\textgreater\ are supported by the evidence stated in \textless description\textgreater\ and the question. Let $F_{\text{ans}}$ denote the set of key items extracted from \textless think\textgreater\,+\,\textless answer\textgreater, and $E$ the items from \textless description\textgreater\,+\,question, defined as
\begin{equation}
\text{cons} \;=\; \frac{|F_{\text{ans}}\cap E|}{\max(1,|F_{\text{ans}}|)},
\qquad
R_{\text{cons}} \;=\;
\begin{cases}
0, & \text{if clear conflicts exist},\\
\text{cons}, & \text{otherwise}.
\end{cases}
\end{equation}
Clear conflicts (e.g., mismatched numbers or attributes) score zero. $R_{\text{cons}}$ rewards alignment between reasoning and perceived evidence, reducing hallucinations and ensuring grounded answers.

%% file: chap/experiments.tex
\section{Experiments}

\subsection{Settings}
\label{sec:set}

\noindent \textbf{Dataset.} 
For SFT, we sample 12K instances from LLaVA-CoT(4k)~\citep{xu2024llava} and Vision-SR1(8k)~\citep{li2025self}, converting each into the structured format \textless description\textgreater, \textless think\textgreater, and \textless answer\textgreater to train the model to verbalize task-relevant visual evidence before reasoning.
For RL, we aggregate multimodal reasoning samples from MMK12(5k)~\citep{meng2025mm}, LLaVA-CoT(5k), Vision-R1-rl(5k)~\citep{huang2025vision},  and Mulberry(5k)~\citep{yao2024mulberry}, spanning domains such as mathematics, science, and figure comprehension. This diverse dataset supports both perception enhancement and general reasoning improvement.

\noindent \textbf{Benchmark.} 
We evaluate multimodal understanding and reasoning using a comprehensive suite: MMMU, MathVista, 
AI2D, EMMA, and Creation-MMBench. 

\emph{MMMU.} targets college-level, multi-discipline reasoning with 11.5K image–text questions across six core disciplines. 

\emph{MathVista.} contains 6{,}141 problems drawn from 28 existing datasets plus three newly curated ones, assessing mathematical reasoning in visual contexts. 

\emph{AI2D.} evaluates diagram understanding on thousands of annotated grade-school science diagrams paired with multiple-choice questions. 

\emph{EMMA.} measures integrated cross-modal reasoning in mathematics, physics, chemistry, and coding, requiring organic image–text reasoning. 

\emph{Creation-MMBench.} specifically assesses context-aware creative capabilities of MLLMs with 765 instances over 51 fine-grained tasks and instance-specific criteria (we also report results on its text-only variant, Creation-MMBench-TO).

\noindent \textbf{Training details}.
For supervised fine-tuning (SFT), we initialized from Qwen2.5-VL-7B-Instruct and tuned all parameters on a merged multimodal dataset of $\sim$12K samples for 3 epochs, with a learning rate of $1\times 10^{-5}$, weight decay of 0.1, batch size 1, and gradient accumulation of 8. bf16 precision, gradient checkpointing, and DeepSpeed ZeRO-3 were enabled to support long-context multimodal inputs.
For reinforcement learning (RL), we adopted a DAPO-style framework implemented in EasyR1-perc, distributed with Ray across one main node and an additional ORM node for reward computation. Tensor parallel size was set to 4. The reward function combined answer accuracy, format compliance, key visual/textual information, n-gram penalty, and consistency, with tuned weights. RL training ran on $\sim$22K samples for 2 epochs.

\subsection{Evaluation}
\label{sec:eval}

\begin{table}[t]
\centering
\setlength{\tabcolsep}{6pt}
\resizebox{1.0\textwidth}{!}{
\begin{tabular}{lcccccc}
\toprule
Method & MathVista & MMMU & EMMA & AI2D & C-MMBench & C-MMBench-TO \\
\midrule
Vision-SR1-7B          & 71.1 & 54.9 & 28.3 & 81.0 & 39.7 & 42.4 \\
Vision-R1-7B           & \textbf{71.3} & 44.9 & 27.4 & 63.2 & 52.6 & 48.5 \\
Perception-R1-7B       & 67.2 & 50.9 & 27.6 & 77.4 & 41.7 & 47.9 \\
Visionary-R1            & 65.5 & 46.2 & 28.2 & 80.5 & 35.1 & 37.6 \\
MM-Eureka-Qwen-7B       & 71.4 & 54.7 & 28.0 & 78.9 & 44.0 & 45.1 \\
VL-Rethinker-7B        & 72.4 & \textbf{56.4} & 27.5 & 79.7 & 41.3 & 45.4 \\
\midrule
Qwen2.5-VL-7B-Instruct & 66.4 & 48.4 & 28.0 & 77.2 & 43.1 & 45.0 \\
\textbf{\modelname} (Before RL)            & 66.4 & 50.6 & 26.6 & 80.4 & 46.7 & 47.0 \\
\textbf{\modelname}           & 71.0 & 52.2 & \textbf{28.8} & \textbf{82.5} & \textbf{53.2} & \textbf{50.5} \\
\bottomrule
\end{tabular}
}
\caption{Performance Comparison Across Multimodal Benchmarks}
\label{tab:final-results}
\end{table}
We report performance across six representative multimodal reasoning benchmarks in Table~\ref{tab:final-results}. Our method (\modelname), after reinforcement learning (RL), consistently outperforms its supervised fine-tuning (SFT) baseline and demonstrates competitive or superior results compared to existing strong baselines.

\textbf{Comparison with Prior Methods.}
\modelname\ achieves new state-of-the-art results on four out of six benchmarks—AI2D (82.5), Creation-MMBench (53.2), C-MMBench-TO (50.5), and EMMA (28.8). Notably, it surpasses VL-Rethinker (79.7) and MM-Eureka (78.9) on AI2D, highlighting its advantage on diagram-heavy perception tasks. On EMMA and both variants of C-MMBench, our model significantly improves over Vision-R1, Perception-R1, and MM-Eureka, demonstrating that incorporating both visual and textual perception rewards yields more reliable and grounded outputs.

\textbf{Effectiveness of Reinforcement Learning.}
The RL stage contributes substantial improvements across all tasks. Compared to the SFT-only version of our model, RL brings gains of +4.6 on MathVista, +1.6 on MMMU, +2.2 on EMMA, +2.0 on AI2D, +6.5 on C-MMBench, and +3.5 on C-MMBench-TO. These gains validate not only the utility of reinforcement learning but also highlight the  contribution of our method: explicitly enhancing the model’s visual and textual perception capabilities. The perception-aware reward design—targeting key visual elements, textual cues, and their consistency—plays a central role, particularly in benchmarks requiring compositional reasoning and fine-grained evidence tracking. The largest improvements on C-MMBench and AI2D support our hypothesis that grounding the \textless description\textgreater\ $\rightarrow$ \textless think\textgreater\ $\rightarrow$ \textless answer\textgreater\ pipeline in concrete perceptual signals leads to more faithful, interpretable, and robust reasoning.

Overall, the consistent improvements across all benchmarks—particularly over perception-focused baselines like Perception-R1, Visionary-R1, and Vision-SR1—demonstrate that \modelname’s integration of explicit perception and structured RL rewards offers a robust and generalizable path forward for multimodal reasoning.

\subsection{Ablation Study}

To demonstrate the effectiveness of our proposed method \methodname, we conducted controlled experiments under different reward configurations. For a fair comparison, all models are initialized from the same SFT checkpoint, trained with 12k samples for 3 epochs, and subsequently fine-tuned on the same RL dataset for one epoch. Performance is evaluated across multiple benchmarks, as shown in Table~\ref{tab:results}.

\begin{table}[t]
\centering
\setlength{\tabcolsep}{6pt}
\resizebox{1.0\textwidth}{!}{
\begin{tabular}{lcccccc}
\toprule
Method & MathVista & MMMU & EMMA & AI2D& C-MMBench & C-MMBench-TO \\
\midrule
Full model (Ours) & 65.0 & 47.9 & \textbf{26.3} & \textbf{80.8} & \textbf{44.5} & \textbf{47.9} \\
\quad - Consistency  & 64.2 & 47.1 & 26.2 & 79.6 & 41.2 & 46.2 \\
\quad - Textual key-info  & 64.3 & \textbf{49.2} & 25.6 & 80.4 & 41.9 & 44.6 \\
\quad - Visual key-info  & \textbf{66.6} & 46.7 & 26.1 & 78.8 & 43.9 & 46.5 \\
\bottomrule
\end{tabular}}
\caption{Ablation results on multiple benchmarks.}
\label{tab:results}
\end{table}

\paragraph{Ablation Analysis.}
The removal of any single reward results in a drop in overall performance, confirming the complementary roles of the three reward components. The consistency reward has the broadest impact: its absence causes the most severe declines on reasoning-intensive benchmarks (-3.26 on  C-MMBench and -1.70 on C-MMBench-TO) and also reduces performance on MathVista and AI2D. This highlights the importance of enforcing a coherent \textless description\textgreater\ $\rightarrow$ \textless think\textgreater\ $\rightarrow$ \textless answer\textgreater\ reasoning chain.

The text perception reward is especially critical for benchmarks like C-MMBench that rely heavily on precise textual cues, where its removal leads to drops of -2.64 and -3.31. Although a slight improvement is observed generally, likely due to reduced over-regularization, most datasets show decreased performance—indicating that weakening textual grounding tends to obscure key constraints.

The visual perception reward contributes most to diagram- or image-intensive tasks, as shown by its impact on AI2D (-2.01) and MMMU (-1.21). Interestingly, its removal slightly improves MathVista (+1.60), suggesting that certain samples may not depend strongly on fine-grained visual grounding.

In summary, the full \methodname\ configuration achieves the best trade-off across all tasks, confirming the necessity of integrating all three reward signals for robust and generalizable performance.

%% file: chap/conclusion.tex
\section{conclusion}
This paper presents \methodname, a unified two-stage training framework that explicitly decouples perception from reasoning in multimodal large language models. Through a comprehensive empirical study across four challenging benchmarks, we demonstrate that explicit perception—especially when paired with textual cues—consistently improves reasoning performance, particularly for smaller models. \methodname\ integrates perception-augmented supervised fine-tuning with perception-aware reinforcement learning, using novel visual, textual, and consistency rewards. Extensive experiments confirm that this perception-grounded approach not only enhances answer accuracy and robustness, but also offers a transparent and auditable path toward more reliable multimodal reasoning. Our results highlight the critical importance of grounding model reasoning in both visual and textual evidence, and pave the way for future work on integrating retrieval, tool use, and knowledge augmentation in perception-aware training pipelines.

%% file: chap/appendix.tex
\section{Appendix}
\subsection{More Preliminary Validation Cases and Framework}
\label{sec:case_ver}

\begin{figure}[htbp]
\includegraphics[width=\linewidth]{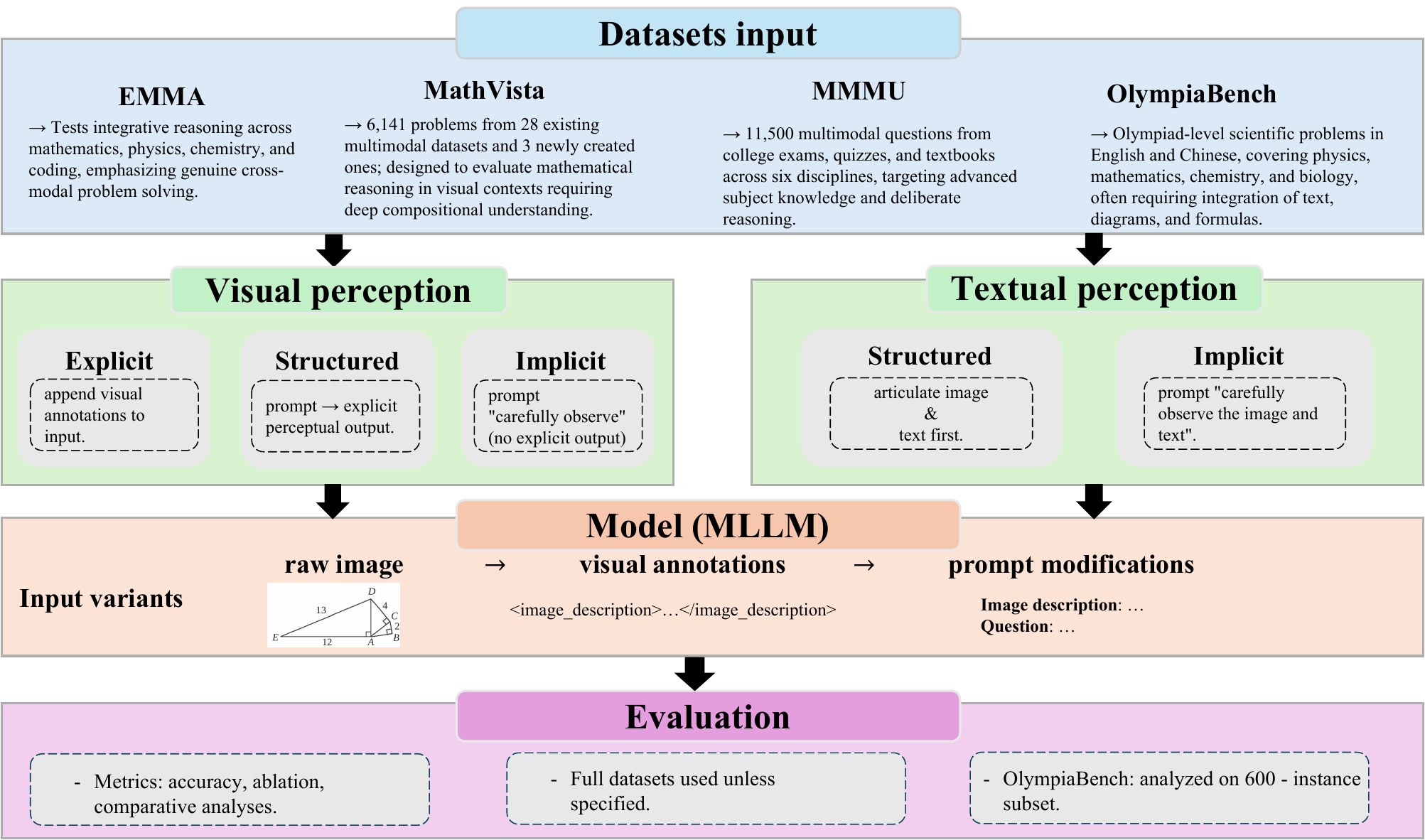}
\caption{Verification experimental pipeline: datasets used, perception conditions (visual and textual), model input variants, and evaluation protocol.}
\label{fig:verification-pipeline}
\end{figure}

\begin{figure}
    \centering
    \includegraphics[width=0.85\linewidth]{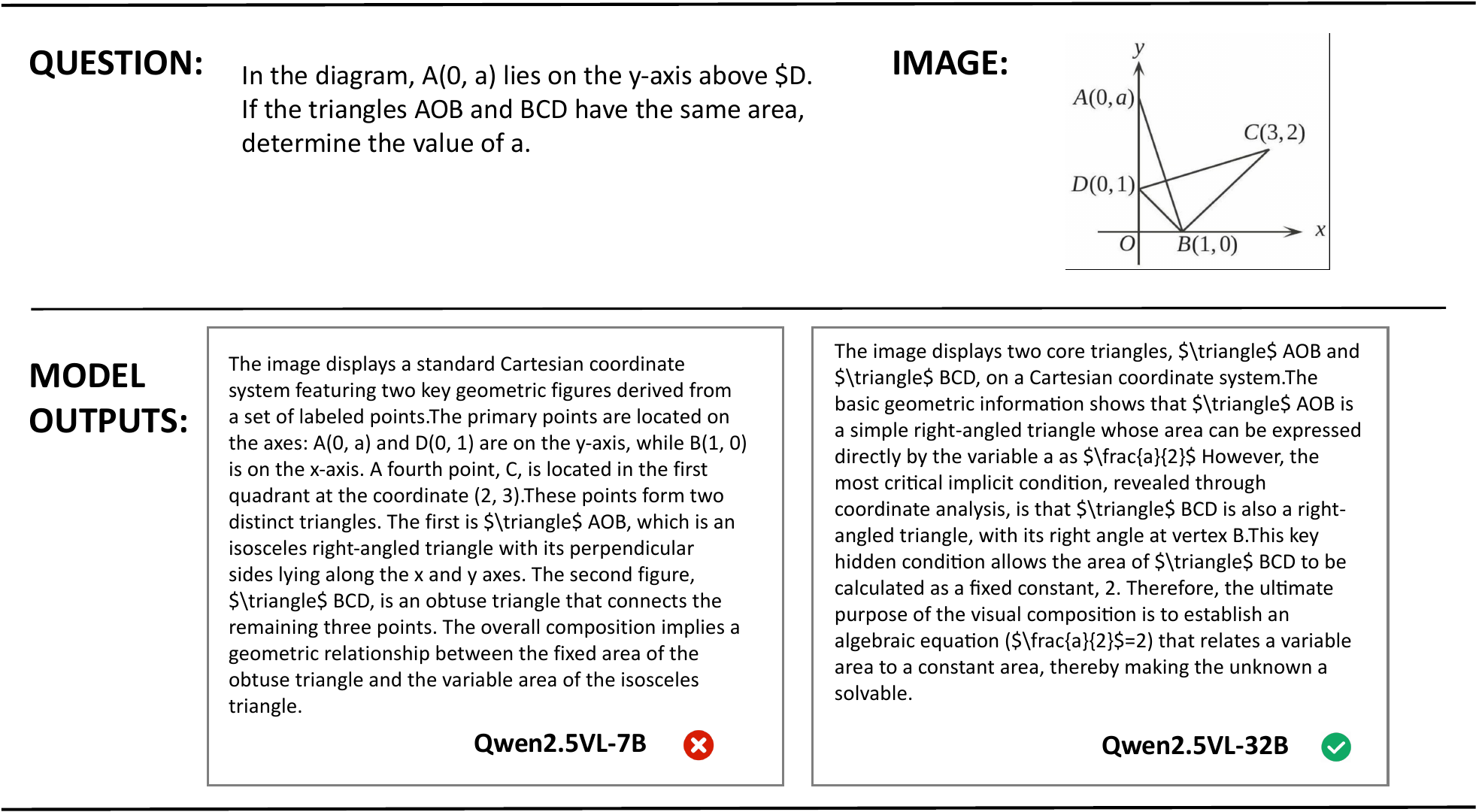}
    \vspace{-0.1in}
    \caption{Case study comparing Qwen2.5-VL-32B and Qwen2.5-VL-7B on a visual grounding task. The larger model correctly identifies $\triangle$AOB as a right-angled triangle, while the smaller model mislabels it as isosceles, illustrating the role of model capacity in perception accuracy.    \label{fig:case_vis}}
\vspace{-0.15in}
\end{figure}

\begin{figure}
    \centering
    \includegraphics[width=1.0\linewidth]{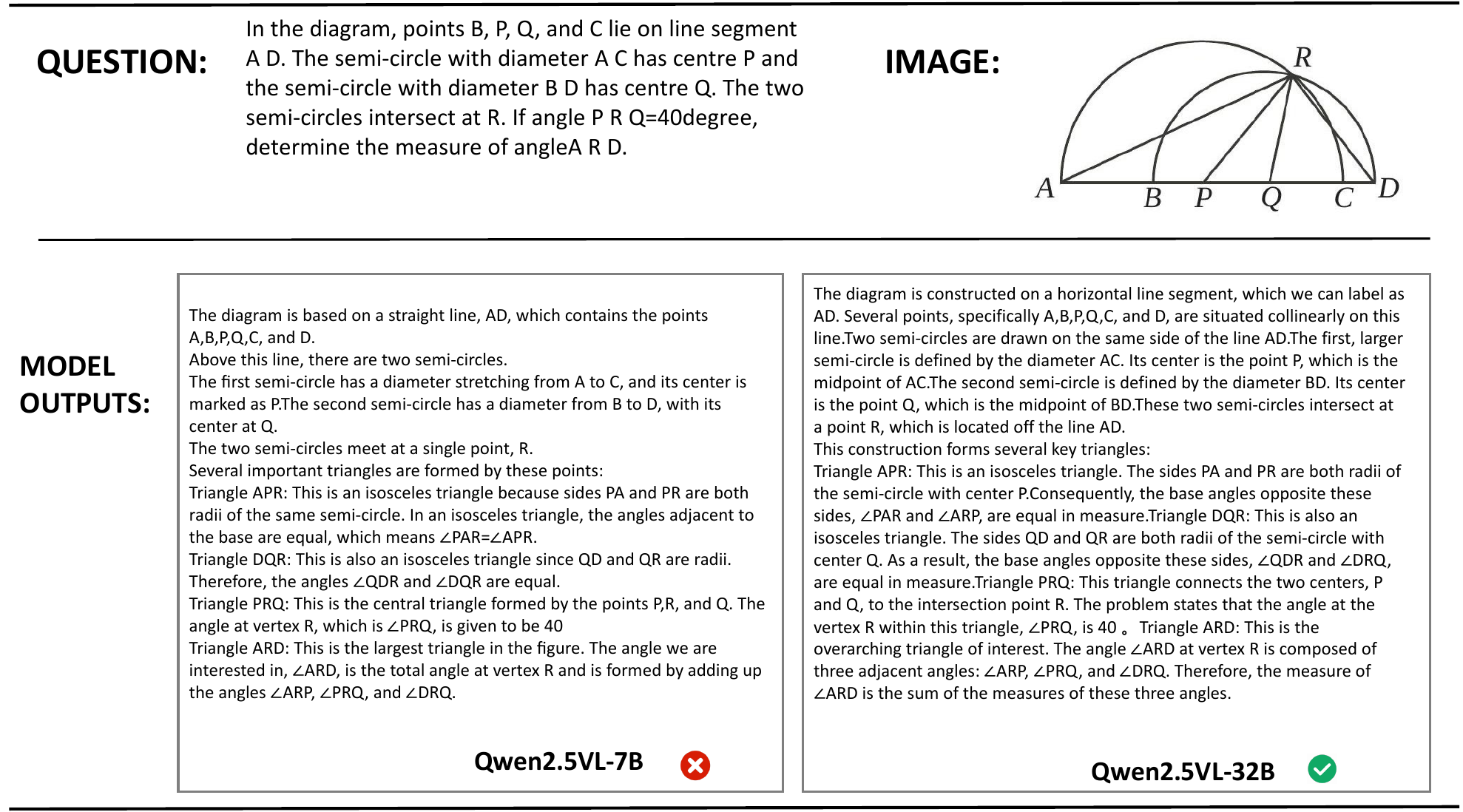}
    \caption{Caption}
    \label{fig:case_visual1}
\end{figure}

\begin{figure}
    \centering
    \includegraphics[width=1.0\linewidth]{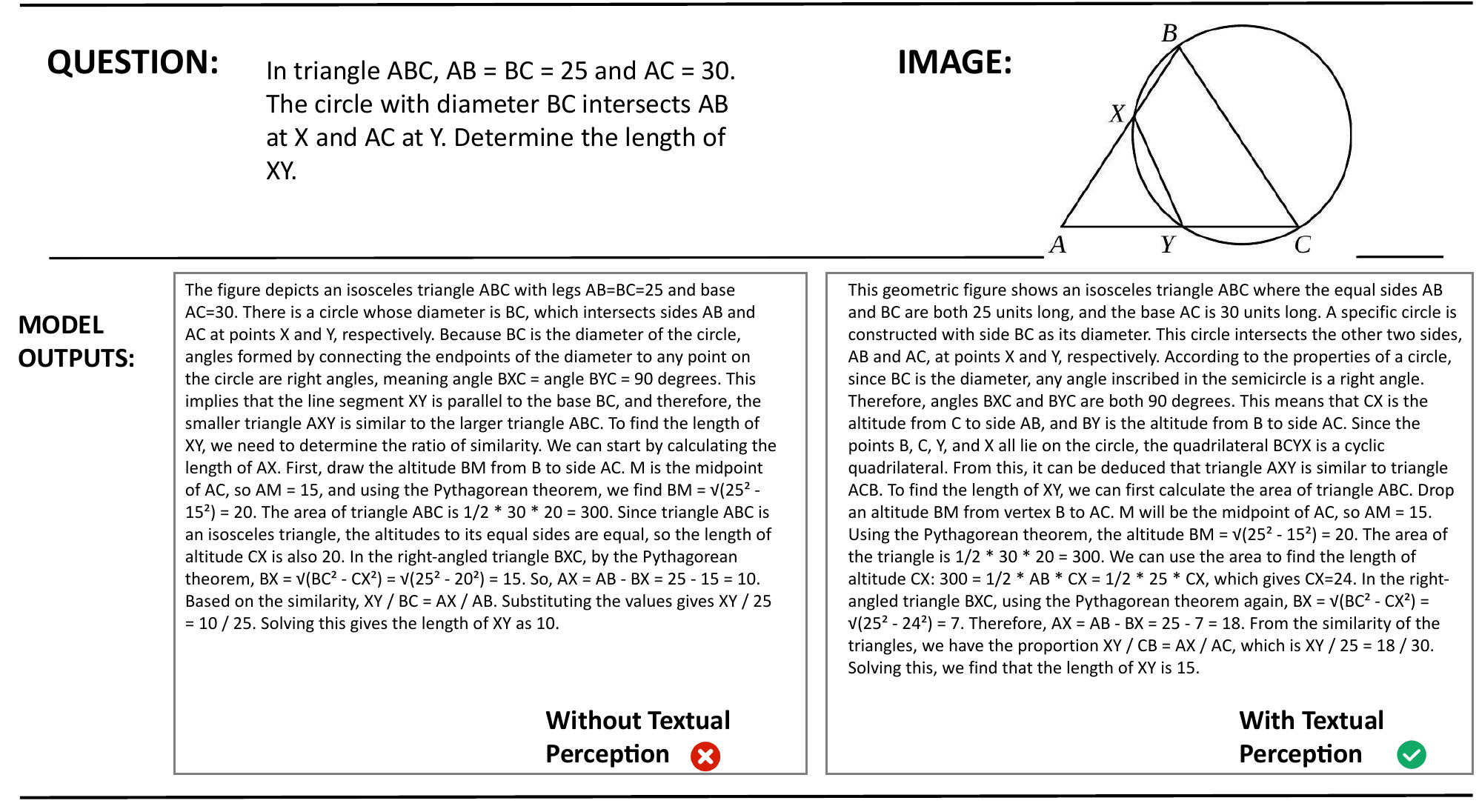}
    \caption{Caption}
    \label{fig:case_text}
\end{figure}

As shown in ~\ref{sec:case_ver}, the performance gap between models of different scales is clearly illustrated by a case study comparing Qwen2.5-VL-32B and Qwen2.5-VL-7B. When prompted for visual grounding, the 32B model showed a solid grasp of the input: it not only captured the obvious features but also inferred hidden geometric properties, correctly identifying $\triangle$AOB as a right-angled triangle—an insight that guided its reasoning effectively. In contrast, the 7B model misinterpreted the diagram, hallucinating an extra property and wrongly describing $\triangle$AOB as an isosceles right-angled triangle. This flawed perception disrupted its reasoning and led to an incorrect solution.

This example provides direct evidence for our central claim: the effectiveness of visual grounding prompts depends on the model’s inherent perceptual ability. For a stronger model, prompting helps unlock deeper understanding and is as effective as providing external annotations. For a weaker model, however, it can introduce errors and degrade performance.

\paragraph{Verification framework (Fig.~\ref{fig:verification-pipeline}\textit{case\_visual1})).}
Figure~\ref{fig:verification-pipeline} summarizes our verification pipeline across datasets, perception conditions, input variants, and the evaluation protocol. We benchmark two MLLMs under three visual settings—\emph{explicit visual notes} (appending curated annotations to inputs), \emph{structured visual grounding} (forcing a description before reasoning), and \emph{implicit visual grounding} (light prompt to “look carefully”)—and two visual–textual settings (structured vs.\ implicit), using full datasets except for a 600-instance, proportionally sampled subset of OlympiaBench. Accuracy is reported with ablations and comparative analyses to isolate the contribution of perception strategies. This setup operationalizes our central question: how explicit versus implicit perceptual grounding modulates downstream reasoning across tasks.  
\paragraph{Visual perception case (Fig.~\ref{fig:case_visual1}, \textit{case\_visual2}).}
In the semicircle–intersection construction, models must decompose the figure into isosceles sub-triangles centered at the semicircle midpoints ($\triangle APR$ and $\triangle DQR$) and compose $\angle ARD$ from base angles and the given $\angle PRQ$. The larger model reliably enumerates the right substructures and completes the angle composition, whereas the smaller model tends to provide verbose but incomplete reasoning. This gap exemplifies our empirical finding: explicit visual notes help both models, but structured prompting can burden weaker perception and lead to degraded solutions—highlighting the capacity dependence of perception prompting observed in our broader study.  

\paragraph{Textual perception case (Fig.~\ref{fig:case_text}, \textit{case\_text}).}
This geometry example illustrates how textual perception stabilizes reasoning. The problem asks for the length of $XY$ in an isosceles triangle $ABC$ with $AB{=}BC{=}25$ and $AC{=}30$, where a circle with diameter $BC$ intersects $AB$ and $AC$ at $X$ and $Y$. Without textual perception, the model introduces spurious assumptions and predicts an incorrect value; with textual perception (making the diameter–right-angle property explicit and driving similarity/power-of-point reasoning), the model recovers the correct answer $XY{=}15$, demonstrating that lightweight textual cues can suppress hallucinations and guide the correct derivation.

\subsection{Dataset Cleaning and Optimization Pipeline for SFT}
\label{app:cleaning-pipeline}

This appendix documents the automated cleaning and optimization pipeline used to produce a high-quality variant of the SFT dataset. The pipeline is modular and deterministic (given fixed random seeds and LLM versions), and it emphasizes (1) converting visual inputs into a single objective representation, (2) re-deriving Chain-of-Thought (CoT) reasoning from that representation, and (3) scoring and selecting samples by multi-dimensional quality metrics.

\subsubsection{Overview}
Given a raw dataset 
\[
\mathcal{D} = \{s_i\}_{i=1}^N, \quad 
s = (\mathrm{id},\,\text{image},\,\text{question},\,\text{original cot},\,\text{original answer})
\]
the pipeline transforms each sample into an enriched record:
\[
\tilde{s} = (\mathrm{id},\; \text{formal\_description},\; \text{cot\_thinking},\; \text{final\_answer},\; \text{quality\_metrics},\; \text{metadata})
\]
Only samples with an overall quality score above a configurable threshold \( \tau \) are retained for the final cleaned dataset; an optional intelligent sampling step can further select a representative subset.

\subsubsection{Step-by-step pipeline}
\textbf{Step 1: Image analysis and formal description generation}

The goal is a single, objective, machine-readable representation of image content (the \emph{formal image description}).
\begin{enumerate}
  \item VLM dense captioning: call a visual-language model (e.g., \texttt{gpt-4o} with image input or an analogous VLM) to generate a dense caption describing layout, objects, and relationships.
  \item Object detection: run Grounding DINO to output object classes and bounding boxes.
  \item OCR: run EasyOCR. The OCR subroutine implements error-handling and fallback strategies to avoid pipeline interruption.
  \item Merging the dense caption, detection outputs and OCR text into a structured canonical text format. This description is the only image-derived input used in later CoT reconstruction to ensure objectivity.
\end{enumerate}

\textbf{Step 2: Chain-of-Thought restructuring}
\begin{itemize}
  \item Input: \texttt{question} and \texttt{formal description}.
  \item Prompt an LLM to produce a new reasoning trace under the explicit instruction: ``Use only the provided formal image description and the question. Do not consult the original CoT or external knowledge beyond the description.''
  \item This re-derivation corrects hallucinations in original CoTs and enforces that reasoning follows observable facts.
\end{itemize}

\textbf{Step 3: Quality assessment and data filtering}

Quality checking uses multi-dimensional metrics (1) does the formal description correctly reflect the image (consistency checks, spot-check prompts to the VLM)? (2) logical clarity and stepwise correctness of reasoning trace. (3) agreement between reasoning trace and final answer. (4) absence of unsupported claims or hallucinations. And a learned/LLM-based scorer.Return per-dimension scores in \([0,1]\). A weighted sum can be used:
    \[
      \text{overall\_score} = w_f s_f + w_c s_c + w_a s_a + w_m s_m,
    \]
    with \(w_f+w_c+w_a+w_m=1\). Values of \(w_\cdot\) are experiment hyperparameters (typical: \(w_f=0.30, w_c=0.35, w_a=0.30, w_m=0.05\)).

\subsubsection{Pipeline pseudocode}
\begin{algorithm}
\caption{SFT Dataset Cleaning \& Optimization Pipeline}
\begin{algorithmic}[1]
\Require raw dataset \(D\), configuration \(C\)
\State Initialize \(\mathcal{P} \leftarrow \varnothing\) (passed set), \(\mathcal{F}\leftarrow\varnothing\) (failed set)
\ForAll{sample \(s\) from \texttt{SFTDataLoader}(D,C)}
  \State \(\text{formal} \leftarrow \text{ImageFormalDescriptionGenerator}(s.\text{image}, s.\text{question}, C)\)
  \State \(\text{cot} \leftarrow \text{CoTRestructurer}(s.\text{question}, \text{formal}, C)\)
  \State \(\text{answer} \leftarrow \text{AnswerExtractor}(\text{cot}, s.\text{question}, C)\)
  \State \(\text{metrics} \leftarrow \text{QualityChecker}(\text{formal}, \text{cot}, \text{answer}, C)\)
  \If{\(\text{metrics.overall\_score} \ge C.\text{min\_score}\)}
    \State add enriched record \(\tilde{s}\) (including metrics) to \(\mathcal{P}\)
  \Else
    \State add \(\tilde{s}\) (marked failed) to \(\mathcal{F}\)
  \EndIf
\EndFor
\If{C.enable\_sampling}
  \State \(\mathcal{S} \leftarrow \text{DataSampler}(\mathcal{P}, C)\)
  \State \Return \(\mathcal{S}\)
\Else
  \State \Return \(\mathcal{P}\)
\EndIf
\end{algorithmic}
\end{algorithm}

\subsubsection{Output schema (example)}
Each processed record is stored as JSON/JSONL with the following example structure:
\begin{verbatim}
{
  "id": "sample_0001",
  "image_path": "images/0001.jpg",
  "question": "What color is the mug on the left?",
  "formal_description": "<structured text listing objects, positions, OCR, etc.>",
  "cot_thinking": "<reconstructed chain-of-thought>",
  "final_answer": "left mug is white",
  "quality_metrics": {
    "formal_score": 0.94,
    "cot_score": 0.88,
    "answer_score": 0.97,
    "overall_score": 0.92
  },
  "passed_quality_check": true,
  "metadata": {
    "detected_objects": [
      {"class":"mug","bbox":[x1,y1,x2,y2]}
    ],
    "ocr_text": "",
    "timestamps": {...}
  }
}
\end{verbatim}
\subsubsection{Metadata fields (tabular summary)}
\begin{table}[h]
\centering
\caption{Key fields stored with each processed sample.}
\begin{tabular}{@{}ll@{}}
\toprule
Field & Description \\
\midrule
\texttt{id} & Unique sample identifier \\
\texttt{formal\_description} & Canonical, objective image description (string) \\
\texttt{cot\_thinking} & Reconstructed chain-of-thought (string) \\
\texttt{final\_answer} & Extracted final answer (normalized string/type) \\
\texttt{quality\_metrics} & Per-dimension scores and \texttt{overall\_score} \\
\texttt{passed\_quality\_check} & Boolean pass/fail \\
\texttt{metadata.detected\_objects} & Object detector output (list) \\
\texttt{metadata.ocr\_text} & OCR-extracted text (string) \\
\texttt{processing\_log} & Warnings, retries, and module traces \\
\bottomrule
\end{tabular}
\end{table}

\paragraph{Concluding remark.} The pipeline converts noisy multimodal CoT data into a high-quality, structured dataset suitable for training and evaluation. The modular design allows swapping or upgrading detectors, OCR backends, and LLMs while preserving the single-source-of-truth principle: all reasoning must be grounded in the generated formal description.

\subsection{Data Construction for RL}
\label{sec:data_rl}
\begin{algorithm}[t]
\caption{Data construction for RL: Visual \& Textual Key Info}
\label{alg:keyinfo-distill}
\begin{algorithmic}[1]
\Require Dataset $D$ of samples $(\mathbf{x}, q, a^\star)$, teacher models $\mathcal{M}$, per-teacher sample count $N$, judge budget $B$, thresholds $(\tau_{\text{acc}}, \tau_{\text{coh}}, \tau_{\text{cons}})$
\Ensure Distilled set $\mathcal{P}$ with CoTs and key information

\State $\mathcal{P} \gets \varnothing$
\ForAll{$( \mathbf{x}, q, a^\star ) \in D$}
    \State Generate $N$ CoT samples per teacher $\to \mathcal{T}$
    \State Select top-$B$ by log-probability
    \State Judge each: compute correctness $s_{\text{acc}}$, coherence $s_{\text{coh}}$
    \State Keep candidates where $s_{\text{acc}} \ge \tau_{\text{acc}}$ and $s_{\text{coh}} \ge \tau_{\text{coh}}$
    \If{$a^\star$ is missing}
        \State Filter by self-consistency
    \Else
        \State Remove duplicates by final answer
    \EndIf
    \State Select best trajectory $\tilde{t}$ from verified candidates
    \State Extract visual key info $\mathcal{V}$ and textual key info $\mathcal{Z}$ from $\tilde{t}$
    \State Add $(\mathbf{x}, q, \hat{a}, \tilde{t}, \{\mathcal{V}, \mathcal{Z}\})$ to $\mathcal{P}$
\EndFor
\State \Return $\mathcal{P}$
\end{algorithmic}
\end{algorithm}

\paragraph{Method Overview.}
Algorithm~\ref{alg:keyinfo-distill} outlines our pipeline for constructing high-quality reinforcement learning (RL) data. It involves three main steps:
(i) sampling diverse chain-of-thought (CoT) trajectories from multiple strong teacher models,
(ii) verifying the quality of these trajectories using a budgeted evaluation process, and
(iii) extracting two types of supervision signals: \emph{visual key info} and \emph{textual key info}.
Each final training instance is represented as a tuple \((\mathbf{x}, q, \hat{a}, \tilde{t}, \{\mathcal{V}, \mathcal{Z}\})\), where $\hat{a}$ is a verified answer, $\tilde{t}$ is a validated trajectory, and $\mathcal{V}, \mathcal{Z}$ are the extracted key information components used in perception-aware RL.

\paragraph{Teacher Ensemble and Diversity.}
For a given input item \(s = (\mathbf{x}, q, a^\star)\), we query several 72B-scale teacher models using stochastic decoding to generate a diverse set of reasoning paths \(\mathcal{T}\). These models produce different but plausible solutions—such as varied reasoning structures, subgoal decompositions, or interpretation strategies—while remaining anchored by accurate prior knowledge.

\paragraph{Budgeted Verification.}
To manage annotation cost efficiently, we first rank candidate trajectories by model log-likelihood (\textsc{TopKByLogProb}) and select the top \(B\) for further evaluation. Each selected trajectory \(t\) is scored based on:
(i) \emph{Correctness} \(s_{\text{acc}}\): measured by exact or normalized answer match, or by tolerance-based comparison if \(a^\star\) is numeric.
(ii) \emph{Coherence} \(s_{\text{coh}}\): assessed by a forward–backward consistency checker that examines step validity, use of evidence, and logical soundness.
Only candidates passing both thresholds \((\tau_{\text{acc}}, \tau_{\text{coh}})\) are retained in \(\mathcal{K}\). When no ground-truth answer is available, we apply a \textsc{SelfConsistencyFilter} to retain trajectories that reach similar conclusions. Otherwise, we de-duplicate based on final answers. We then select a single best trajectory \(\tilde{t}\) using a weighted score: \(w_1 s_{\text{acc}} + w_2 s_{\text{coh}}\).

\paragraph{Key Info Extraction.}
From the verified trajectory \(\tilde{t}\), we extract two complementary types of supervision:
\begin{itemize}
    \item \textbf{Visual key info} \(\mathcal{V}\): This includes atomic, image-grounded facts such as object attributes, visual measurements, geometric constraints (e.g., parallel lines, equal lengths), and spatial relationships. These are precise and verifiable visual elements crucial for grounded reasoning.
    \item \textbf{Textual key info} \(\mathcal{Z}\): This includes (i) a structured parse of the question's textual facts—such as entities, quantities, conditions, and units—and (ii) an \emph{application map} that links these facts to specific reasoning steps in \(\tilde{t}\), explicitly showing how each fact is used during problem solving. This provides a fine-grained, executable understanding of the question content.
\end{itemize}

\paragraph{Output and Usage in RL.}
Each training item stores the verified answer \(\hat{a}\), trajectory \(\tilde{t}\), and extracted key information \(\mathcal{V}, \mathcal{Z}\). These are used during RL to define perception-level rewards by checking how well the model’s outputs align with the key information. These are then combined with answer correctness and format-based rewards. The combination of budgeted verification and self-consistency ensures high-quality data while controlling annotation cost.

\paragraph{Differences from Prior Visual-Only Pipelines.}
Unlike previous pipelines that rely solely on visual supervision, our method:
(i) unifies both \emph{visual} and \emph{textual} perception signals,
(ii) introduces a budget-aware selection strategy that reduces unnecessary annotation by filtering weak candidates early, and
(iii) formalizes textual key info not just as extracted facts, but as a structured fact-to-reasoning mapping, making question constraints explicit and testable.